\documentclass[runningheads]{llncs}
\usepackage{amsmath}

\usepackage[T1]{fontenc}
\usepackage{graphicx}

\usepackage{hyperref}
\usepackage{textcomp}
\usepackage{gensymb}
\usepackage{arydshln}
\usepackage{booktabs}
\usepackage{accvabbrv}
\usepackage{amssymb}
\newcommand{\aggr}{\operatorname{aggr}}
\usepackage[dvipsnames]{xcolor}

\newcommand{\green}[1]{{\color{Green}#1}}

\usepackage{pdfpages}

\newcommand\blfootnote[1]{%
  \begingroup
  \renewcommand\thefootnote{}\footnote{#1}%
  \addtocounter{footnote}{-1}%
  \endgroup
}

\usepackage{color}

\begin{document}

\title{Leveraging Semantic Cues from Foundation \\ Vision Models for Enhanced Local \\ Feature Correspondence}
\titlerunning{Leveraging Semantic Cues for Local Feature Correspondence}

\author{Felipe Cadar\inst{1,2} \and
Guilherme Potje\inst{1} \and 
Renato Martins\inst{2,3} \and 
Cédric Demonceaux\inst{2,3} \and 
Erickson R. Nascimento\inst{1,4}}

\authorrunning{F. Cadar et al.}
%
\institute{Computer Science Department, Universidade Federal de Minas Gerais, Brazil \\ 
\email{\{cadar, guipotje, erickson\}@dcc.ufmg.br}
\and
ICB UMR CNRS 6303, Université de Bourgogne, France \\
\email{\{renato.martins,cedric.demonceaux\}@u-bourgogne.fr}
\and
Inria, LORIA, CNRS, Université de Lorraine, France \and
Microsoft \\ 
}
\maketitle              

\begin{abstract}
Visual correspondence is a crucial step in key computer vision tasks, including camera localization, image registration, and structure from motion.
The most effective techniques for matching keypoints currently involve using learned sparse or dense matchers, which need pairs of images. These neural networks have a good general understanding of features from both images, but they often struggle to match points from different semantic areas. This paper presents a new method that uses semantic cues from foundation vision model features (like DINOv2) to enhance local feature matching by incorporating semantic reasoning into existing descriptors. Therefore, the learned descriptors do not require image pairs at inference time,
allowing feature caching and fast matching using similarity search, unlike learned matchers.
We present adapted versions of six existing descriptors, with an average increase in performance of $29\%$ in camera localization, with comparable accuracy to existing matchers as LightGlue and LoFTR in two existing benchmarks. Both code and trained models are available at 
\url{https://www.verlab.dcc.ufmg.br/descriptors/reasoning_accv24/} \blfootnote{This work has been accepted to the 17th Asian Conference on Computer Vision (ACCV 2024)}

\end{abstract}
\section{Introduction}
\label{sec:intro}

Visual correspondence is fundamental for important higher-level vision tasks like camera pose estimation, simultaneous localization and mapping (SLAM), and structure from motion (SfM). Recently, the pipeline for finding visual correspondences between pairs of images has been changing in favor of methods that provide different types of context aggregation, like learned sparse matchers \cite{cit:superglue,cit:lightglue} or dense correspondence networks \cite{cit:loftr}. These methods depend on gathering information from both perspectives to condition features for better correspondence prediction. Although they have been shown to provide better results in downstream tasks, they must be run for every pair of images, making it expensive to use in large tasks like SfM pipelines, where a single image will be matched multiple times to other images with a similar viewpoint.
While the traditional single-view pipeline can pre-extract features for individual images and use an efficient similarity search such as mutual nearest neighbor (MNN), it does not perform as well as the context aggregation methods. This paper proposes an approach to semantically condition keypoint descriptors to find better and more consistent correspondences while maintaining the advantages of single-view extraction and caching.
\begin{figure}[tb]
  \centering
  \includegraphics[width=0.95\columnwidth]{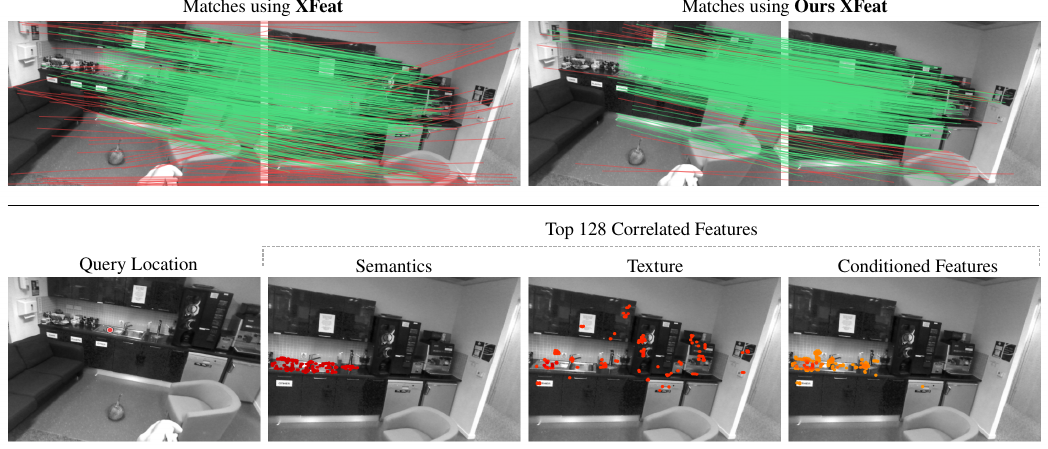}
  \caption{\textbf{Leveraging semantic information for improving visual correspondence.} The figure illustrates the matching process using Mutual Nearest Neighbor (MNN) for the base descriptor XFeat~\cite{cit:xfeat} and for our approach, which employs semantic conditioning (shown in the top right). Correct matches are shown in green and wrong matches in red. We can also assess the interpretability and consistency of the descriptors by finding the closest $128$ matches to a given query point in the image (red point in the bottom left) using either semantic or texture features. Hotter colors indicate higher similarities. Please notice the similarity ranking improvement with the conditioned features around the sink region.}
  \label{fig:teaser}
\end{figure}
Foundations Models, such as DINOv2~\cite{cit:dinov2} and SAM~\cite{cit:sam}, can extract features that contain an understanding of semantic concepts in a scene to complement local texture patterns. These features can be adapted for various tasks, \textit{e.g.}, image classification, instance retrieval, video understanding, depth estimation, semantic segmentation, and semantic matching, by freezing the backbone and training new layers for a specific task \cite{cit:dinov2}. In order to capture the meaning of scenes and objects, models such as DINOv2 have developed a strong invariance to local texture changes. However, the high level of invariance in these features can make them less sensitive when it comes to identifying pixel-level matches between images. Instead, they can offer a basis for agreement between regions, which can be used to filter connections between visually similar but semantically different regions, as shown in Fig.~\ref{fig:teaser}. In this paper, instead of relying on two-view context aggregation, we propose an effective technique to leverage high-level features understanding coming from an LVM to semantically condition textured-based correspondences. 

A key technical contribution of this paper is a novel learning-based method for integrating semantic context into local features, enabling efficient similarity search during matching and substantially improving matching accuracy. The experimental results demonstrate that our approach significantly enhances the performance of various detect-and-describe techniques in camera pose estimation and visual localization tasks within indoor environments.

\section{Related Works}
\label{sec:related}

\subsubsection{Semantics and object-level in description.}

Leveraging semantic features for their high invariance is not unprecedented. In nonrigid image matching, methods like GeoPatch~\cite{cit:geopatch}, DEAL~\cite{cit:deal}, and DALF~\cite{cit:dalf} merge geometric and texture features to achieve invariance to deformations. Sim2Real~\cite{cit:sim2real} is a detector and describe approach that learns to find correspondences using two types of losses: the inter-objects, and the intra-objects, segmenting its features into two parts. The first part is responsible for finding the right object, and the second part matches a precise location within that object. Although it captures the concept of separating invariance and distinctiveness, its features are trained specifically for single object-matching. Our approach introduces a much more general formulation, significantly improving camera pose estimation in indoor scenes. SFD2~\cite{cit:sfd2} also introduced a more general approach by explicitly distilling semantic segmentation features. To improve long-term outdoor visual localization, it uses the semantic cues from an off-the-shelf semantic segmentation model to classify keypoints in four levels of stability, including {\it Volatile}, {\it Dynamic}, {\it Short-term}, and {\it Long-term}. The descriptor learning uses the same semantic backbone as the detector, and similar to Sim2Real, it optimizes both intra and inter-class distances. Although SFD2 proposes a more general matching descriptor, it still fixes the classes of keypoints that are reliable and depends on fixed semantic segmentation labels for learning descriptors. Our proposed network is different since we do not explicitly define any semantic behavior based on classes, making it more general and easier to train while also leveraging foundation vision models, which provides a fine-grained and more general semantic understanding of the scene. DeDoDe descriptor~\cite{cit:dedode} is the closest competitor to our methodology and optimizes a dual-softmax matching loss while incorporating DINOv2~\cite{cit:dinov2} features in the extraction pipeline. Although it uses the same semantic cues as our methodology, DeDoDe descriptors do not perform any aggregation between features. In our proposed model, we design an attention mechanism to refine the description based on the semantic cues of the scenes, in addition to learning a semantic-only feature vector for improving the matching.

\subsubsection{Context aggregation.}
A recent trend in image correspondence for finding more robust matchings between views is to use information from both views to predict an assignment. The learned sparse matchers SuperGlue~\cite{cit:superglue} and more recently LightGlue~\cite{cit:lightglue} receive keypoints and descriptors from two images from a base extractor method, like SuperPoint~\cite{cit:superpoint} and predict an assignment of the keypoints. Using self and cross-attention, these networks iteratively refine the descriptors into ``matching descriptors''. In this process, every descriptor is aware of the other descriptors from its own image and from the corresponding pair view. The motivation is that the attention aggregation process can gather enough context information to estimate a more consistent and robust matching while discarding keypoints that are not visible in both images. Recently, Omniglue sparse matcher~\cite{cit:omniglue} extended LightGlue by adding semantic features from DINOv2 into the pipeline, still maintaining the two-view aggregation. 

Methods like LoFTR~\cite{cit:loftr} and EfficientLoFTR~\cite{cit:eloftr}
adopt linear approximations of the attention aggregation and a coarse-to-fine strategy to predict an assignment, respectively. In the same direction, MESA~\cite{cit:mesa} also uses a detection-free matching pipeline, but including higher-level features from the Segment Anything Model~\cite{cit:sam}. Overall, methods that aggregate context from both views, either dense or sparse, cannot fully cache features between images. 

This aggregation strategy in large-scale reconstruction can be expensive since the number of image pairs to be matched can grow quadratically. Our proposed method, on its turn, only uses a single view for  the feature extraction and aggregation, allowing frameworks to cache features for later use in an efficient similarity search operation.

\section{Methodology}
\label{sec:method}

\begin{figure}[tb]
  \centering
  \includegraphics[width=0.99\columnwidth]{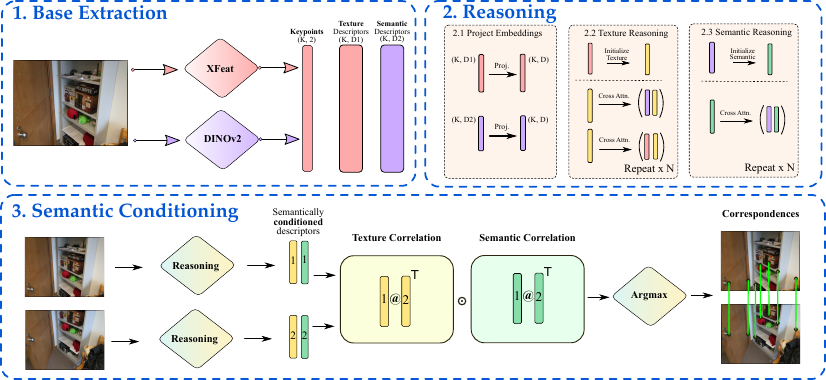}
  \caption{\textbf{Semantic Conditioning Pipeline.} Our method first extracts both low-level scene texture via a base local feature descriptor (XFeat) and semantically meaningful, high-level features via a foundation vision model (DINOv2) with the associated salient texture keypoints. Then, the Reasoning module is applied in both representations, where cross attention layers are used iteratively to enhance the representations of both texture and semantic features. Finally, the descriptor similarity is computed by combining both the texture and semantic similarity using element-wise product ($\odot$).}
  \label{fig:methodology}
\end{figure}

In this section, we present the main concept of our methodology, detailing how to add semantic awareness to a local descriptor and the supervision designed to train it. A scheme of the overall training and inference stages is shown in Fig.~\ref{fig:methodology}. The proposed strategy first extracts two sets of descriptors: a set of \textbf{texture} features using an off-the-shelf local feature method, and the \textbf{semantic} features coming from a LVM model for context (as the selected DINOv2 in this work). To that end, we adopt a base method for extracting traditional, textured-focused features and a base method for extracting semantic-focused features. Following the base extraction, we refine the features using a self-attention \textit{Reasoning} module. For finding matching image pairs, we use the two sets of \textbf{texture} and \textbf{semantic} features, extracted independently for each image, to calculate a similarity matrix for finding mutual matches using the \textit{Semantic Conditioning}. 

\subsubsection{Base Feature Extraction.}
\label{sec:base-features-extraction}
We extract two sets of base features: the texture-based ($D^*_{t}$) and the semantic-based ($D^*_{s}$) descriptors. For extracting texture-based descriptors and keypoints, we use an off-the-shelf detection and description method, like XFeat~\cite{cit:xfeat} or Superpoint~\cite{cit:superpoint}. Following, we extract a dense feature map from DINOv2 and sample the semantic-based descriptors at the positions of the detected keypoints via bicubic interpolation. Finally, we project both sets of descriptors to a shared dimension size,
as shown in the top left side of Fig. \ref{fig:methodology} (\emph{Base Extraction}).

\subsubsection{Descriptor Reasoning.}
\label{sec:reasoning}

At this point, using the two sets of features (\emph{i.e.}, texture-based and semantic-based) we  refine them for the \textit{Semantic Conditioning} stage. We initialize the refined texture descriptor ($D_t$) and the refined semantic descriptor ($D_s$) using their raw versions $D^*_{s}$ and $D^*_{t}$. 
We iteratively refine $D_t$ using attention aggregation while alternating the attention keys between the raw descriptors $D^*_{s}$ and $D^*_{t}$. For each attention layer $\aggr_i(K,Q,V)$, we compute:
\begin{equation}
  \begin{aligned}
    D_{t_{i+1}} = \aggr_i( D^*_{\textbf{s}}, D_{t_{i}}, D_{t_{i}} )  & \ \ \  \text{if } i \text{ is even}, \\
    D_{t_{i+1}} = \aggr_i( D^*_{\textbf{t}}, D_{t_{i}}, D_{t_{i}} )  & \ \ \ \text{if } i \text{ is odd},
  \end{aligned}
\end{equation}
\noindent where $i$ is the number of the iteration, $K$ the key, $Q$ the query, $V$ the value, and $D_{t_{i}}$ and $D_{t_{i+1}}$ are the refined texture descriptors at the iterations $i$ and $i+1$. The attention aggregation mechanism is the same as the Lightglue's cross-attention layer \cite{cit:lightglue}. Similarly, we iteratively refine $D_s$, using only the original semantic features as keys:
\begin{equation}
  \begin{aligned}
    D_{s_{i+1}} = \aggr_i( D^*_{\textbf{s}}, D_{s_{i}}, D_{s_{i}} ).
  \end{aligned}
\end{equation}
The strategy of using alternating texture and semantic keys aims to incorporate both types of information, providing the texture descriptor with scene context from the semantic descriptors. This approach enables the texture component to identify precise matches and to understand the type of structure it is describing, resulting in more semantically relevant aggregated information. Similarly, we do not want the refined semantic features to have a local understanding of the scene. The objective of this reasoning pipeline is to guide the matches for semantically coherent areas. Finally, both sets of descriptors are normalized to the unit hyper-sphere with the $L_2$ normalization.

\subsubsection{Semantic Conditioning.}
\label{sec:semantic-conditioning}

After obtaining $D_t$, $D_s$, and the base extractor's keypoints $K$, we cache them all for later use since there are no more steps before the visual matching. The extraction from a pair of images $I^1$ and $I^2$ will result in $D^1_t$, $D^1_s$, $K^1$ (descriptors and keypoints for the first image), and $D^2_t$, $D^2_s$, $K^2$ (descriptors and keypoints for the second image). With each pair of descriptor types, we can calculate a correlation matrix $C_t$ and $C_s$, for texture and semantics, respectively. Individually, $D_t$ and $D_s$ are either too local or too global, resulting in discriminative descriptors that can wrongly match repetitive textures or highly invariant descriptors that can only find coarse matches. We define the {\it Semantic Conditioning} as the operation of combining both correlation matrices for adjusting the match probability between coherent regions while keeping the local discriminatory power. The final correlation matrix considering the product of features is then
\begin{equation}
  \begin{aligned}
    C_f = C_t \odot C_s, 
  \end{aligned}
\end{equation}
\noindent where $\odot$ is the element-wise product operation. $C_f$ is then used for finding matches with Mutual Nearest Neighbor (MNN) search. We conditioned the matching using the product operation so that even low semantic similarities can completely erase wrong matches, regardless of high texture similarity.

\subsection{Supervision}
\label{sec:supervision}

For training, the parameters of the base extractors remain frozen and we optimize only the weights of the initial projections and the descriptor reasoning as depicted in Fig.~\ref{fig:methodology}.
We freeze the weights because each base extractor can have a specific training strategy that works better. By using freezed, off-the-shelf extractors, we can accommodate more methods. The DINOv2 was also frozen following \cite{cit:dinov2}, which used it as a backbone for several tasks.

We trained the networks using the ScanNet \cite{cit:scannet} dataset, which contains exclusively indoor scenes. We focused on indoor environments because they are more semantically rich than outdoor scenes. We sample image pairs with a minimum estimated overlap of $40\%$, along with the depth maps and camera poses.
For every pair, we use the base extractor to estimate keypoints from both images, project the keypoints from the first one to the second image, and find matches that are under three pixels distance from a correspondent keypoint. These ground-truth assignments are obtained for calculating the dual-softmax loss, following a similar strategy to XFeat~\cite{cit:xfeat}. Since our descriptors are refined iteratively, we can follow the supervision process of LightGlue~\cite{cit:lightglue} and calculate the dual-softmax loss for each iteration of the descriptors to seep up convergence. Assuming the ground-truth assignment $M_{gt}$ of size $M\times2$ containing the matching indices, we calculate the loss $\mathcal{L}_l$ for the layer $l$ as:
\begin{equation}
  \begin{aligned}
    \mathcal{L}_{l} = \ & - \sum_{(i,j) \in M_{gt}} \log(\text{softmax}_r (C_{f_l})_{(i,j)}) \\ 
              & - \sum_{(i,j) \in M_{gt}} \log(\text{softmax}_r (C^{\top}_{f_l})_{(j,i)}), 
  \end{aligned}
\end{equation}
\noindent where $\text{softmax}_r$ is a row-wise softmax function and the correlation matrix $C_{f_l}$ for the layer $l$ is computed as:

\begin{equation}
  \begin{aligned}
    C_{t_l} = D^1_{t_{l}}D^{2\top}_{t_{l}}, \quad
    C_{s_l} = D^1_{s_{l}}D^{2\top}_{s_{l}}, \quad
    C_{f_l} = C_{t_l} \odot C_{s_l}.
  \end{aligned}
\end{equation}

Following the deep supervision of LightGlue, the final loss is the mean of the intermediary layers' losses. For $N$ layers, the final loss $\mathcal{L}_{\text{total}}$ is then: 
\begin{equation}
  \begin{aligned}
    \mathcal{L}_{\text{total}} =\frac{1}{N}{\sum_{l \in [1,N]} \mathcal{L}_{l}}.
  \end{aligned}
\end{equation}

\section{Experiments}
\label{sec:experiments}

\subsubsection{Experimental setup and training.}
We trained the method in approximately $1{,}000{,}000$ pairs from the ScanNet dataset~\cite{cit:scannet}, with a fixed image size of $512\times512$, extracting $2{,}048$ keypoints and descriptors on each image.
The training takes $10$ hours on $4\times$V100 32GB GPUs with a batch size of $16$ on each GPU.
We used ${5}$ layers of attention aggregation. The experiment with $9$ layers used a batch size of $8$ but kept the same total training paris. We used Adam with learning rate $1e^{-4}$ as the optimizer and the descriptor size is fixed to $256$. We have adopted DINOv2 as the semantics provider as it has been shown to provide rich semantic features that generalize well in different benchmarks~\cite{cit:dedode,cit:omniglue,cit:dinov2}.

For the DINOv2 version, we tested all four types and settled on the smallest one (S), as discussed in Sec.~\ref{sec:ablation}.
For the evaluation, we have adopted the Scannet1500 relative pose estimation benchmark, which inclides $1{,}500$ image pairs from the test split of ScanNet, and the 7Scenes \cite{cit:7scenes} visual localization benchmark, which contains $7$ indoor scenes. Both benchmarks pose challenges for the visual correspondence task, as they contain motion blur, repetitive structures, and low-textured surfaces.

\subsubsection{Inference and computational analysis.}
In inference time, we resize input images to $896$ pixels on the longest edge for extracting DINOv2 features, but we keep the original shape for the base extractor. For all models, we normalize the output descriptors. All experiments were run without resizing the images. For a pair of images in the original ScanNet resolution of $968\times1296$, and with a max keypoint detection count of $2{,}048$, our network takes, on average, 23ms for texture features extraction, 83ms for semantic features extraction, 4ms for reasoning, and 11ms for matching, using XFeat as the base local extractor, for a total of 121ms, in which 110ms are processed in single images and can be cached. All experiments were executed in a NVIDIA GeForce RTX 3080 with 10GB.

\begin{table}[t!]
  \caption{\textbf{Pose estimation results compared to global visual matchers and existing single view competitors}. The semantic conditioning yields results that are even competitive with learned matchers and the detector-free matcher LoFTR (half upper part of the table), as well as to existing single view descriptors with semantics (our direct competitors shown in the bottom part).}
  \label{tab:scannet1500_small}
  \centering
      \resizebox{0.4\textwidth}{!}{
      \begin{tabular}{lr@{\hskip 0.15in}r@{\hskip 0.15in}r}
    \toprule
        & \multicolumn{3}{c}{AUC $\uparrow$ } \\  \cline{2-4} 
     Method      & @5\degree & @10\degree \ & @20\degree \\ \midrule
     LoFTR          & 20.20 & 37.60 & 52.60  \\ 
     LightGlue & 22.30 & 40.80 & 57.00  \\  \hdashline
     DeDoDe-G & 9.50 & 19.40 & 30.20  \\
     SFD2 & 13.00 & 25.60 & 38.30  \\
     Ours ~~~~   &  \textbf{20.30} & \textbf{36.81} & \textbf{52.37} \\ 
  \bottomrule
  \end{tabular}}
\end{table}

\begin{table}[t!]
  \caption{\textbf{Semantic conditioning capability of our method for different base descriptors}. ``Ours'' accompanying a method's name means that we trained our network with the features of this base method. Pose estimation results on Scannet1500. }
  \label{tab:scannet1500}
  \centering
      \resizebox{0.55\textwidth}{!}{\begin{tabular}{l@{\hskip 0.15in}r@{\hskip 0.15in}r@{\hskip 0.15in}rc}
    \toprule
        & \multicolumn{3}{c}{AUC $\uparrow$ } & Average  \\  \cline{2-4} 
     Method      & @5\degree & @10\degree \ & @20\degree & Gain \\ \midrule
     DeDoDe-B & 5.70 & 11.30 & 17.80 &  \\
     Ours+DeDoDe-B  & \textbf{8.91} & \textbf{17.28} & \textbf{26.25} & \green{+50.67\%} \\\midrule
     DeDoDe-G & 9.50 & 19.40 & 30.20 &  \\
     Ours+DeDoDe-G  & \textbf{11.13} & \textbf{21.29} & \textbf{32.04} & \green{+9.07\%} \\ \midrule
     ALIKE & 7.60 & 15.40 & 23.90 &  \\
     ALIKE Ours & \textbf{10.56} & \textbf{20.96} & \textbf{31.92} & \green{+35.27\%} \\ \midrule
     ALIKED & 12.00 & 23.40 & 35.80 &  \\
     Ours+ALIKED  & \textbf{15.31} & \textbf{29.39} & \textbf{43.85} & \green{+24.37\%} \\ \midrule
     RELF  & 11.60 & 24.20 & 33.30 &  \\
     Ours+RELF   &  \textbf{15.20} & \textbf{30.46} & \textbf{46.31} &  \green{+24.12\%} 
     \\ \midrule
     XFeat & 15.80 & 30.80 & 45.80 & \\
     Ours+XFeat   & \textbf{19.72} & \textbf{36.77} & \textbf{53.19} &  \green{+18.70\%}  \\ \midrule
     SuperPoint  & 13.70 & 26.30 & 39.90 &  \\
     Ours+SuperPoint   &  \textbf{20.30} & \textbf{36.81} & \textbf{52.37} &  \green{+37.02\%} \\ 
  \bottomrule
  \end{tabular}}
\end{table}

\subsubsection{Baselines.}

In the selected baselines, we have traditional detect and describe methods as ALIKE~\cite{cit:alike}, ALIKED~\cite{cit:aliked}, SuperPoint~\cite{cit:superpoint}, RELF~\cite{cet:relf} and XFeat~\cite{cit:xfeat}. We selected DeDoDe~\cite{cit:dedode}, as the only other descriptor that uses semantic information from DINO, and SFD2 \cite{cit:sfd2} that explicitly uses semantic information for supervision in training. Although our proposed method is not a matcher, we also included results for the learned matcher LightGlue~\cite{cit:lightglue} and the detector-free matcher LoFTR~\cite{cit:loftr} for comparing methods that can aggregate information from both views.

\subsection{Relative Pose Estimation}
\subsubsection{Setup.} Following \cite{cit:loftr}, we evaluated pose estimation in 1500 pairs of images from the validation split of the ScanNet \cite{cit:scannet} dataset. The indoor images contain strong viewpoint changes, poorly textured surfaces, and motion blur artifacts, making it challenging for the pose estimation task. As in previous works~\cite{cit:xfeat}, we use LO-RANSAC~\cite{cit:lo-ransac} to estimate the essential matrix and search for the optimal threshold for each method. 
\subsubsection{Metrics.} We follow the protocol of previous works~\cite{cit:xfeat,cit:lightglue,cit:loftr}, which report the area under the curve (AUC) of the maximum angular error in rotation and translation at the thresholds 5°, 10°, and 20°.
\subsubsection{Results.} 

The quantitative registration results are shown in Tab.~\ref{tab:scannet1500_small} and Tab.~\ref{tab:scannet1500}. In Table \ref{tab:scannet1500_small}, we included methods of dual-view context aggregation, like LightGlue \cite{cit:lightglue} and LoFTR \cite{cit:loftr}, and other descriptors that also leverage semantic information like DeDoDe-G \cite{cit:dedode} and SFD2 \cite{cit:sfd2}. It is worth noting that even doing only single view extraction, Superpoint combined with semantic conditioning can yield competitive results compared to Lightglue (without any awareness of the pairing view). Our designed strategy to leverage semantics to improve the matching capability of existing descriptors is described in~Tab.~\ref{tab:scannet1500}. We can notice a significant improvement in all baselines when combined with our proposed semantic conditioning. Although many of them were only trained on outdoor images from MegaDepth dataset~\cite{cit:megadepth}, like DeDoDe~\cite{cit:dedode}, SFD2 \cite{cit:sfd2}, ALIKE~\cite{cit:alike}, and ALIKED~\cite{cit:aliked}, we could still improve their indoor pose estimation results by at least $24\%$ without retraining neither the feature extraction nor the DINOv2 backbone. This result suggests that the extracted visual cues are not better than the original versions of these descriptors, only better conditioned by the semantics. 

\subsection{Visual Localization}
\subsubsection{Setup.} We evaluated the visual localization task on the popular benchmark 7Scenes~\cite{cit:7scenes}. The 7Scenes dataset comprises seven different indoor scenes with annotated 6-DoF camera poses and dense 3D reconstructions. It is divided into training and test sequences, each with a distinct camera path. The images are acquired using a Kinect 1 and are often blurry, with several regions featuring repetitive indoor structures and textureless areas. We use the HLoc~\cite{cit:hloc} framework for building a Structure-from-Motion (SfM)~\cite{cit:colmap} database from the training sequences. Then, the images in the test set are localized by computing image correspondences between the current frame and the SfM models using the descriptors provided by each competitor.

\subsubsection{Metrics.}
As a standard practice~\cite{cit:hloc}, for each localized image from the test set, we use the camera translation error in meters, and rotation errors in degrees, according to the ground-truth camera poses provided by the benchmark~\cite{cit:7scenes}.

\subsubsection{Results.} The visual localization benchmark results are shown in Tab.~\ref{tab:7scenes}. An interesting observation is that our method is able to reduce XFeat errors in several cases. For SuperPoint, our approach was not able to provide meaningful improvements. We hypothesize that XFeat being a smaller backbone, provides more concise and less redundant features, making it less prone to overfitting and leverages the semantic information the most.
 We achieve the highest average percentage of correctly localized cameras when considering the percentage of localized cameras within different thresholds. From stricter thresholds of $1^\circ$, $1$ cm up to $500$ cm, $10^\circ$, LightGlue, the gold standard but expensive matcher, correctly localizes $66.97\%$ of the cameras. This is followed by Ours (with SuperPoint as base texture detector): $66.95\%$, Superpoint: $66.88\%$, XFeat: $66.36\%$, and DeDoDe-G: $64.44\%$. This demonstrates that semantic information can increase correspondences in ambiguous regions, as shown in Fig.~\ref{fig:qualitative}.

\begin{table}[t]
  \caption{\textbf{Visual localization results in 7Scenes.} For each scene we report the camera pose error achieved by the competing methods. We also include LightGlue as a learned matcher reference, providing an upper bound in matching quality.}
  \label{tab:7scenes}
  \centering
  \resizebox{\textwidth}{!}{\begin{tabular}{l:r@{,}r:r@{,}r:r@{,}r:r@{,}r:r@{,}r:r@{,}r:r@{,}r}
    \toprule
     & \multicolumn{14}{c}{ Translation errors (m) $\downarrow$, Rotation errors (\degree) $\downarrow$} \\
     Methods $\downarrow$ & \multicolumn{2}{c}{Chess} & \multicolumn{2}{c}{Fire} & \multicolumn{2}{c}{Heads} & \multicolumn{2}{c}{Office} & \multicolumn{2}{c}{Pumpkin} & \multicolumn{2}{c}{Kitchen} & \multicolumn{2}{c}{Stairs} \\ \midrule
     LightGlue & 0.024 & 0.803 & 0.019 & 0.792 & 0.011 & 0.715 & 0.027 & 0.826 & 0.039 & 1.044 & 0.033 & 1.127 & 0.051 & 1.324 \\  \hdashline
    DeDoDe-G & \textbf{0.025} & 0.902 & \textbf{0.018} & 0.793 & \textbf{0.011} & 0.810 & 0.028 & 0.892 & 0.046 & 1.357 & 0.035 & 1.250 & 0.069 & 1.708 \\
    XFeat & 0.027 & 0.919 & \textbf{0.018} & 0.731 & 0.013 & 0.865 & 0.028 & 0.842 & 0.042 & 1.072 & 0.035 & 1.189 & \textbf{0.043} & \textbf{1.108} \\
    Ours+XFeat & 0.026 & 0.913 & \textbf{0.018} & \textbf{0.725} & 0.013 & 0.881 & \textbf{0.027} & 0.840 & 0.043 & 1.111 & 0.035 & 1.215 & 0.048 & \textbf{1.108} \\
    SuperPoint & \textbf{0.025} & \textbf{0.822} & \textbf{0.018} & 0.738 & \textbf{0.011} & \textbf{0.759} & \textbf{0.027} & \textbf{0.820} & \textbf{0.040} & \textbf{1.052} & \textbf{0.034} & \textbf{1.130} & 0.055 & 1.427 \\
    Ours+SuperPoint & 0.026 & 0.876 & 0.019 & 0.755 & 0.012 & 0.802 & \textbf{0.027} & 0.826 & 0.041 & \textbf{1.052} & \textbf{0.034} & 1.171 & 0.046 & 1.167 \\ 
  \bottomrule
  \end{tabular}}
\end{table}

\subsection{Sensitivity and Ablation Studies}
\label{sec:ablation}
In this section, we investigate the design decisions that led to our final semantic condoning method. Every variant in these experiments was evaluated using relative pose estimation on the Scannet1500 benchmark, using XFeat~\cite{cit:xfeat} as the base extractor. The choice of XFeat in these experiments was guided since it is a recent lightweight descriptor with competitive performance along different visual correspondence benchmarks.

\subsubsection{DINOv2 backbone size.}
As the DINOv2 model family has many versions with different capacities, they also require increasingly higher amount of computation. Following DeDoDe~\cite{cit:dedode}, we fixed the dimension of the input images to DINOv2 to mitigate the fast-growing 
computational complexity of the vision transformer, which is quadratic in the number of tokens. To study which type of model and image resolution, we trained four versions of our network, one for each DINO model, and evaluated fixing the input DINOv2 image size in inference to $518$ and $896$ on the long edge. Table~\ref{tab:ablation-dino} shows that, while fixing image size to $518$, larger DINOv2 models can yield better pose estimation results; the results difference is smaller for the larger images. We chose to fix the input size to $896$ and use the smaller version of DINOv2 as it already yields satisfactory results. This also leads to a lower inference time.

\begin{table}[t]
  \caption{\textbf{Sensitivity analysis to semantic model capacity}. We evaluate fixing DINO interval resizing dimension and varying model size.}
  \label{tab:ablation-dino}
  \centering
   \resizebox{0.7\textwidth}{!}{\begin{tabular}{cr@{\hskip 0.15in}r@{\hskip 0.15in}rrc@{\hskip 0.15in}r@{\hskip 0.15in}r@{\hskip 0.15in}rr}
    \toprule
     DINOv2   & \multicolumn{3}{c}{AUC $\uparrow$ @ size 518 } & Inference & & \multicolumn{3}{c}{AUC $\uparrow$ @ size 896 } & Inference \\  \cline{2-4} \cline{7-9} 
    Size      & @5\degree & @10\degree & @20\degree & \multicolumn{1}{c}{time (ms)} & & @5\degree & @10\degree & @20\degree & \multicolumn{1}{c}{time (ms)} \\ \midrule
    S & 18.80 & 36.33 & 53.11 & 54.80  &  & 19.72 & 36.77 & 53.19 & 121.11 \\
    B & 19.87 & 37.57 & 53.88 & 88.86  & & 20.70 & 37.93 & 54.30 & 225.96 \\
    L & 19.54 & 37.13 & 53.95 & 174.54 & & 19.88 & 37.70 & 55.00 & 561.63 \\
    G & 20.41 & 38.20 & 55.22 & 446.01 & & 20.89 & 38.58 & 55.66 & 1695.00 \\
  \bottomrule
  \end{tabular}}
\end{table}

\subsubsection{Semantic conditioning.}

When refining the texture descriptors ($D_t$) with the semantic keys ($D^*_s)$, they can also aggregate semantic information. On Experiment 1 of Tab.~\ref{tab:ablation}, we tested the capacity of a single descriptor playing both roles of semantic and texture guidance. We removed the semantic refinement and only used the texture correlation ($C_t$) for matching descriptors. We can see a large drop in performance in all thresholds compared to Experiment 3, which is our default setting.

\begin{table}[t]
  \caption{\textbf{Pose estimation evaluation on Scannet1500}. SC means Semantic Conditioning, \#AL is the number of aggregation layers, and K is the type of Key used in the texture reasoning. The version used in all experiments is marked in \textbf{bold}.}
  \label{tab:ablation}
  \centering
    \resizebox{0.4\textwidth}{!}{\begin{tabular}{cccccr@{\hskip 0.15in}r@{\hskip 0.15in}r}
    \toprule
       & & & & & \multicolumn{3}{c}{AUC $\uparrow$ } \\  \cline{5-7} 
      Exp & SC &  K & \#AL & @5\degree & @10\degree \ & @20\degree \\ \midrule
      1 &           & A & 5 & 18.27 & 34.29 & 49.80 \\
      2 & \checkmark & A & 3 & 18.71 & 35.57 & 51.68 \\ 
      \textbf{3} & \checkmark & \textbf{A} & \textbf{5} & \textbf{19.72} & \textbf{36.77} & \textbf{53.19} \\ 
      4 & \checkmark & A & 7 & 19.19 & 36.12 & 52.21 \\
      5 & \checkmark  & A & 9 & 19.35 & 36.34 & 52.66 \\ \hdashline
      6 & \checkmark  & S & 5 & 19.18 & 36.19 & 52.60 \\
      7 & \checkmark  & T & 5 & 19.48 & 36.64 & 52.58 \\
  \bottomrule
  \end{tabular}}
\end{table}

\subsubsection{Number of attention aggregation layers.}

Another factor that we investigated was the number of layers for attention aggregation. This evaluation follows an observation from LightGlue~\cite{cit:lightglue}, where the network can collect information from both images to decide if the descriptors are already good for matching or need more refining. In single-view extraction we do not want to depend on other image descriptors, so it is difficult to know when to stop refining dynamically. For that reason, we decided to fix the number of aggregation layers. In Tab.~\ref{tab:ablation}, Experiments 2 through 5 evaluate networks trained with $3$, $5$, $7$, and $9$ layers. Although they all are significantly close to each other, the biggest gap was between $3$ and $5$ layers; for that reason, we kept the five layers as default. 

\subsubsection{Alternating aggregation key.}

We also evaluated the impact of alternating keys in the texture descriptor refinement. In Tab.~\ref{tab:ablation}, experiments 6 and 7 are versions of the default network that only use the semantic key and the texture key, respectively. We can notice that using only the semantic key is worse than using only the texture key. Since the network still has semantic conditioning in the matching stage, it should leverage the texture information for fine matching. Although the texture only is better than semantic only, alternating between both yields better results because the network can adjust the texture features while aggregating context from the semantics of the scene present in the image.

\subsection{Qualitative Matching Analysis}
\begin{figure}[t]
  \centering
  \includegraphics[width=0.95\linewidth]{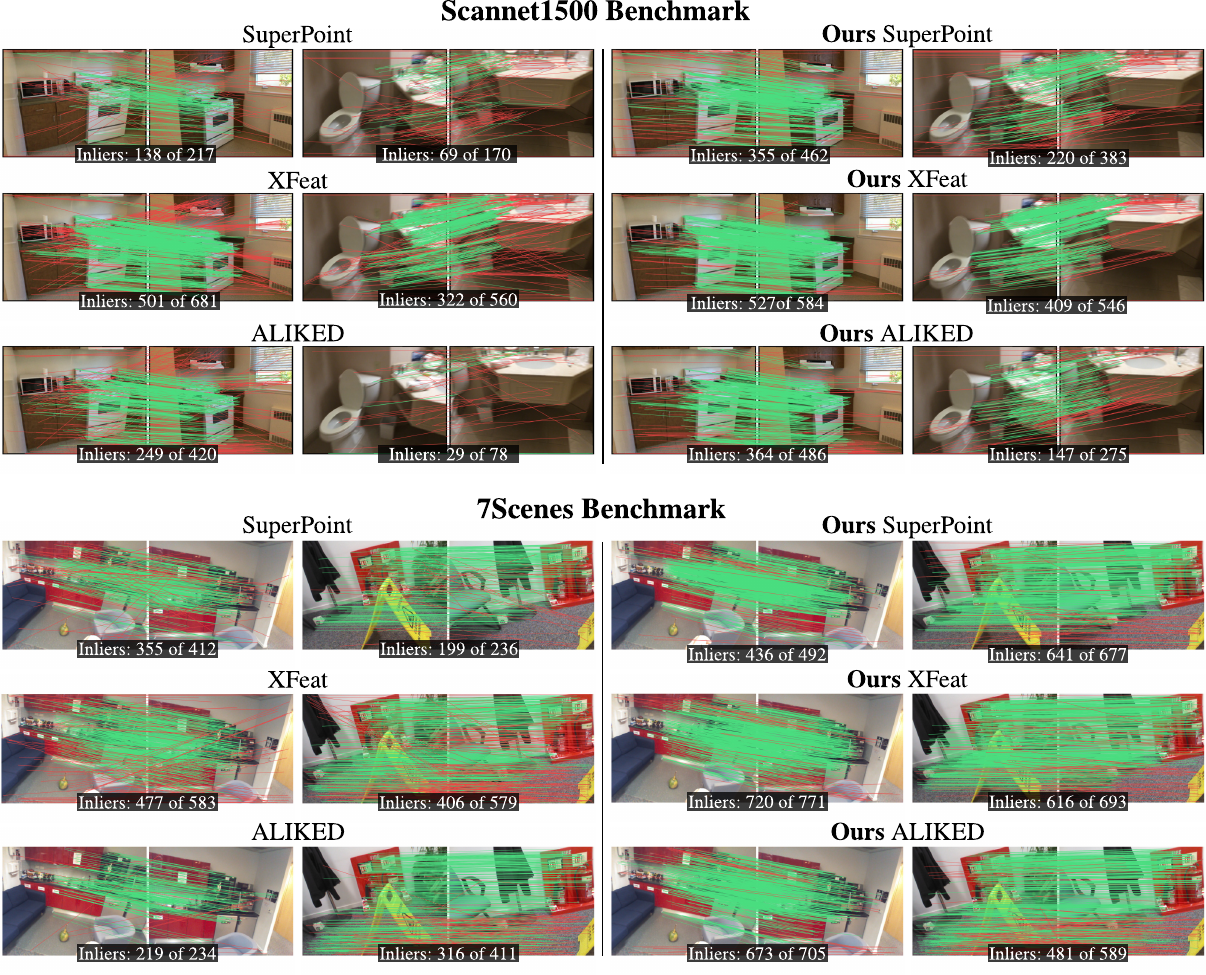}
  \caption{\textbf{Visual matching results for three refined base descriptors in the two benchmarks}. Green matches are inliers and red outlier matches. The left side of the figure shows the results for the original descriptors SuperPoint~\cite{cit:superpoint}, XFeat~\cite{cit:xfeat} and ALIKED~\cite{cit:aliked} when matching two different image pairs. The right side of the figure shows the matching using their semantically conditioned versions using our proposed methodology. We can observe from visual inspection that the matches are more consistent between the views. Please also notice the increased inlier ratio when considering the semantic conditioned versions.}
  \label{fig:qualitative}
\end{figure}

The reason for using semantic cues to find matches is to reduce ambiguity between descriptors from semantically different areas by using both sets of information. This makes it less likely to mistakenly match a similar texture to the wrong instance. As can be noticed in Fig.~\ref{fig:qualitative} with examples from the two considered benchmarks, and without any filtering, our method leverages the semantic information for a much more consistent matching between the views. By using semantical conditioning, we not only filter semantically wrong matches, but we find a greater number of correct correspondences between coherent areas.

\subsubsection{Interpretability of the features.} 

\begin{figure}[t!]
  \centering
  \includegraphics[width=\columnwidth]{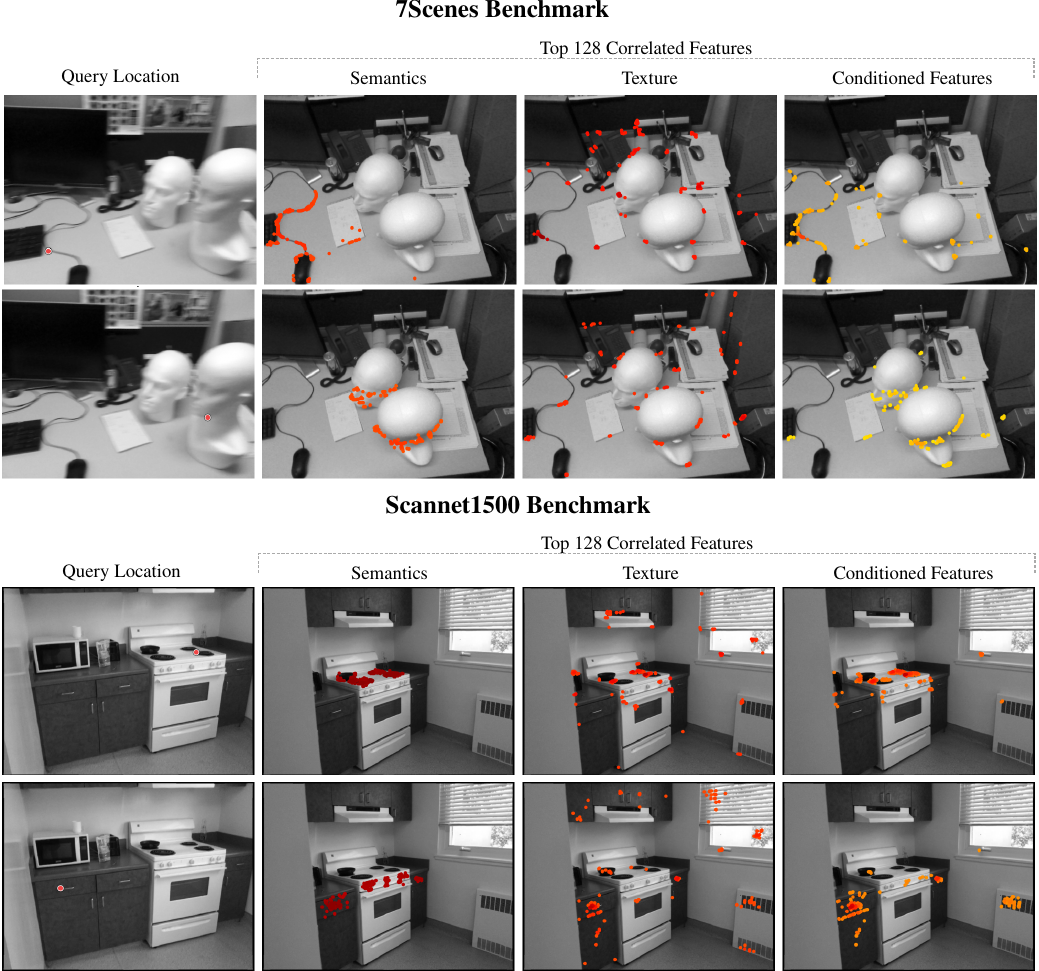}
  \caption{\textbf{Interpretability and consistency of the conditioned features.} We show the closest $128$ matches to a given query keypoint (red point in the first column) for the different descriptors with either solely semantics, the refined texture descriptor or with the proposed semantic conditioned features (fourth column). Hotter colors indicate higher similarity. Please notice the similarity ranking improvement with the conditioned features for finding matches such as in the mouse cable (first row). Our approach consistency is highlighted in the estimated closest keypoints when selecting the drawer handle of the kitchen (fourth row) which is occluded in the paired view.}
  \label{fig:interpretability}
\end{figure}

We can also check the interpretability and consistency of the descriptors by computing the closest 128 matches to a given query point in the image using either solely the semantic, the texture or the semantically conditioned features as shown in Figs.~\ref{fig:teaser} and \ref{fig:interpretability}. Given a query keypoint (highlight in red in the first column) we find the top 128 corresponding keypoints ranked by feature correlation. The semantic correlation selects only keypoints in a similar area, but without local precision. This can be seen in Fig.~\ref{fig:teaser} where the areas belonging to the sink are selected as the closest ones. The same behavior is displayed in the top example in Fig.~\ref{fig:interpretability}, where the query point located in the cable also activates points located along the entire cable. On the other hand, the texture description correlation selects keypoints locally similar but has not regard for the context of the location (as shown in the third column in Figs.~\ref{fig:teaser} and \ref{fig:interpretability}). The conditioned features can find correspondences that are relevant both in semantics and texture. In some cases, we can have high texture and semantic correlations for non corresponding keypoints, for instance when the correct correspondence is not visible, as shown in the last row of Fig.~\ref{fig:interpretability} where the selected keypoint is not visible in the pairing view. Yet, the closest matches in this case are the ones located in the two other handles. Other representative visualization examples of the correlations for four other scenes are provided in the supplementary material.

\section{Conclusion}
\label{sec:conclusion}

This work introduces a learning-based technique for visual feature description that is capable of utilizing semantic cues present in the image. We design a network performing information aggregation that leverages semantic features to refine and condition off-the-shelf descriptors to improve indoor visual matching. Our method performance in camera pose estimation is superior to existing state-of-the-art models exploring semantic cues, and it is also competitive even with that of recent learned matchers (such as LightGlue), while only using a single image for feature extraction and Nearest Neighbor search for matching. 
With extensive experiments, we show that our method can improve the pose estimation results of six different base descriptors by $25\%$ on average. The improved descriptors can be used in large-scale SfM reconstructions by utilizing a single view for image extraction, as MNN matching is much faster than running learned matchers for thousands of image pairs.

\subsubsection{\ackname} This work was partially supported by grants from CAPES, CNPq, FAPEMIG, Google, ANER MOVIS from Conseil Régional BFC and ANR (ANR-23-CE23-0003-01), to whom we are grateful. This project was also provided with AI computing and storage resources by GENCI at IDRIS thanks to the grant 2024-AD011015289 on the supercomputer Jean Zay's V100 partitions.

\bibliographystyle{splncs04}
\bibliography{main}



\section*{Supplementary material}

# [Supplementary Material]
# Leveraging Semantic Cues from Foundation Vision Models for Enhanced Local Feature Correspondence

Felipe Cadar[*1,2], Guilherme Potje[1], Renato Martins[2,3], Cédric Demonceaux[2,3], and Erickson R. Nascimento[1,4]

[1] Computer Science Department, Universidade Federal de Minas Gerais, Brazil
[2] ICB UMR CNRS 6303, Université de Bourgogne, France
[3] Inria, LORIA, CNRS, Université de Lorraine, France
[4] Microsoft

Project page with code and trained models:
https://www.verlab.dcc.ufmg.br/descriptors/reasoning_accv24/

In this supplementary material to the main paper, we provide additional interpretability results of the semantic conditioned features, complementing the qualitative results presented in the paper, more timing statistics, and a deeper analysis on DINO backbone size selection.

## 1 Inference time analysis and deep-matchers

**Table 1. Single Pair Time Analysis**. Although we have a longer total time, we can cache the extraction results and reuse them later.

|  | Time in milliseconds | | |
| --- | --- | --- | --- |
|  | Extraction | Matching | Total |
| Ours | 127.183 | 1.389 | 128.571 |
| LightGlue | 35.381 | 26.674 | 50.657 |
|  | Pairs per second | | |
| Ours | 7.86 | 720.03 | 7.78 |
| LightGlue | 28.26 | 37.49 | 16.11 |

For having a better view of the pratical use of our method we included two timing studies:

**1)** We calculated the extraction and matching time for one pair of images with the ScanNet resolution of $968 \times 1,296$ and 2,048 keypoints. In an NVIDIA GeForce RTX 3080/10GB, LightGlue run the feature extraction in 35ms and matching in 27ms, while our method takes 127ms for feature extraction and 1.3ms for feature matching, as shown in Tab. 1

**2)** To simulate a more realistic scenario, like an SfM pipeline where one image will be matched multiple times, we calculated and cached all keypoints and features for $N$ images and then matched all pairs of images, *i.e.*, $\frac{N(N-1)}{2}$, as usually done in COLMAP. Fig. 1 shows how LightGlue's matching time quickly escalates. Our method becomes more efficient than the deep matcher after just 15 images (a small size dataset).

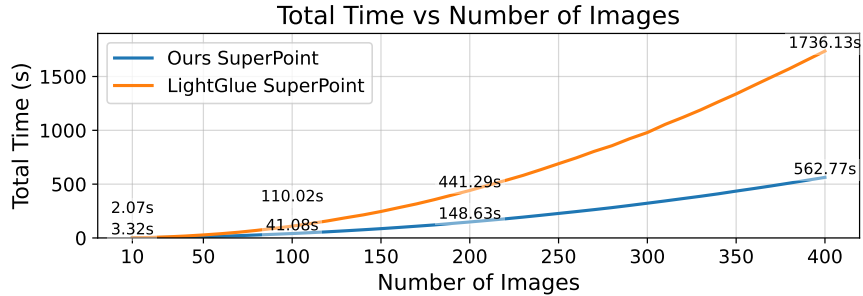

**Fig. 1. Time analysis.** Time for all images keypoints and descriptors and match all pairs of images. The total time includes GPU-CPU transfers and caching the extractions operations.

We also calculated the compute times to run the 7Scenes benchmark. The timing include some other operations performed by the HLoc framework (commonly used to run the benchmark) that may not be optimized, reducing the gap between the methods. Fig. 2 shows the results.

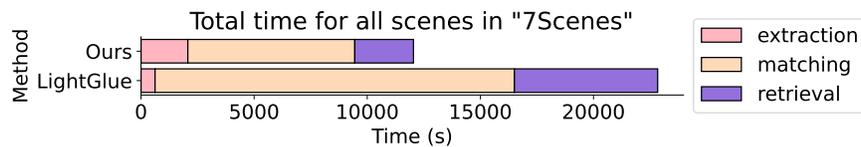

**Fig. 2. Time on the 7Scenes benchmark.**

## 2   Further analysis on DINOv2 backbone size.

The sensitivity study on the DINOv2 backbone size reveals that the larger foundation models do not significantly increase pose estimation performance. It is worth noting that ViTs, as used in DINO, are not suitable for fine matching due to the low spatial resolution of their feature maps, but they have strong semantic understanding capabilities due to their global receptive field and attention mechanism. The goal of using foundation models in our pipeline is to provide semantic guidance for the texture features, and the semantic understanding of

small models have shown to be good in our experiments, while reducing computational overhead. This aligns with results reported in the DINOv2 paper, where tasks requiring general semantic understanding such as instance-level recognition and semantic segmentation (DINOv2 paper – Tabs. 7,9,10), exhibit similar performance across all backbone sizes.

## 3   Interpretability of the features

The proposed description strategy with semantic conditioning also displays interpretable features, which further explain the obtained matching results both quantitatively and from visual inspection (as shown in Fig. 3 of the main paper). We provide more examples of feature interpretability in the following Figs. 3, 4 and 5 of this supplementary material.

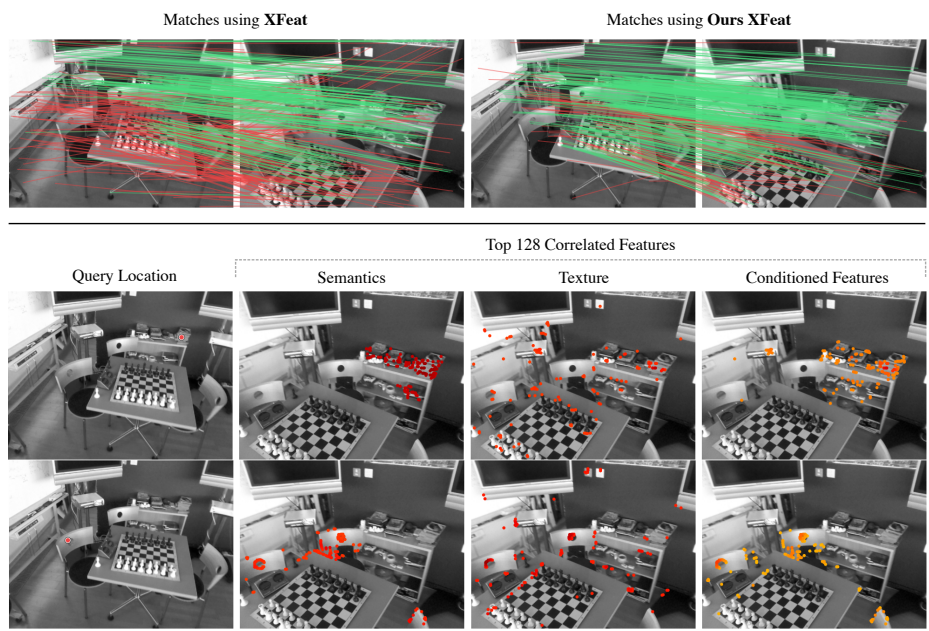

**Fig. 3. Interpretability and consistency of the conditioned features.** We show the closest 128 matches to a given query keypoint (red point in the first column) for the different descriptors with either sole semantics, the refined texture descriptor, or the proposed semantic conditioned features (fourth column). Hotter colors indicate higher similarity. Notice how the semantic information focuses on instances with similar meaning, like the objects on the table and the chairs, guiding the textural features.

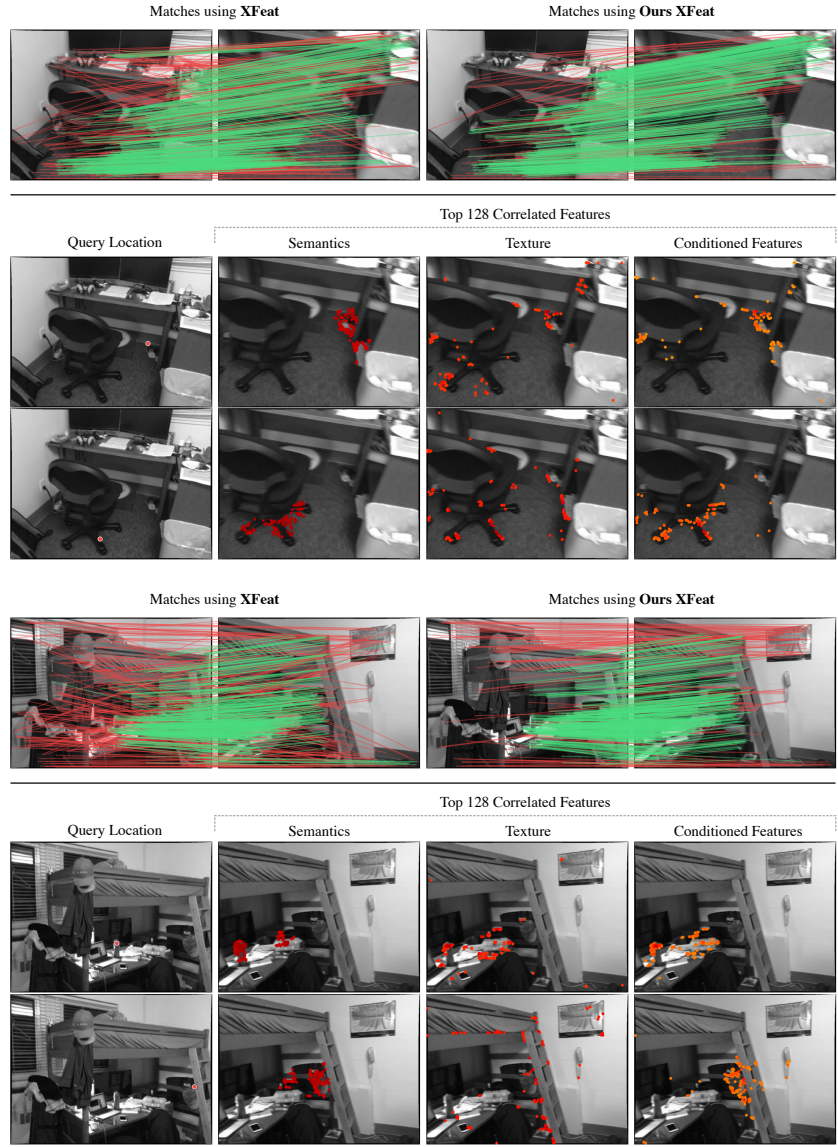

**Fig. 4. Interpretability and consistency of the conditioned features.** In some cases, the original texture-only features can already show a good matching performance, but they still yield many outliers. In these two examples, the semantic information plays a bigger part in filtering lowering the probability of matching areas with different contexts, also lowering the number of outlier matches.

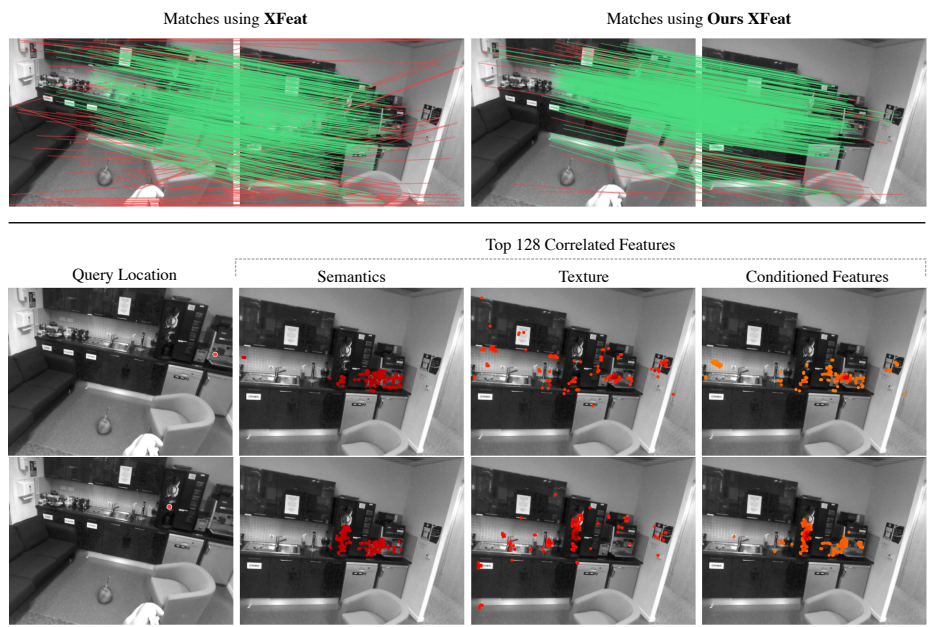

**Fig. 5. Interpretability and consistency of the conditioned features.** Observe how, with similar semantics in both query points, the texture information helps to differentiate the semantics.



\end{document}